\documentclass[a4paper,twoside]{article}

\usepackage{epsfig}
\usepackage{epstopdf}
\usepackage{subcaption}
\usepackage{calc}
\usepackage{amssymb}
\usepackage{amstext}
\usepackage{amsmath}
\usepackage{amsthm}
\usepackage{multicol}
\usepackage{pslatex}
\usepackage{apalike}

\usepackage{url}
\usepackage{graphics}
\usepackage{graphicx}
\usepackage{booktabs}
\usepackage{algorithm}
\usepackage{algorithmic}

\usepackage{SCITEPRESS}

\begin{document}

\title{FrequentNet: A Novel Interpretable Deep Learning Model for Image Classification}

\author{
    \authorname{
        Yifei Li\sup{1}, 
        Kuangyan Song\sup{2},
        Yiming Sun\sup{3}\thanks{Corresponding author.},
        and Liao Zhu\sup{4}\orcidAuthor{0000-0001-7261-1773}
    }
    \affiliation{\sup{1}Department of Computer Science, Zhejiang University, China}
    \affiliation{\sup{2}Microsoft}
    \affiliation{\sup{3}Instacart}
    \affiliation{\sup{4}Department of Statistics and Data Science, Cornell University, Ithaca, USA}
    \email{
        lyeeef@gmail.com,
        nachtsky@gmail.com,
        sunstat.faust@gmail.com,
        lz384@cornell.edu
    }
}


\abstract{This paper has proposed a new baseline deep learning model of more benefits for image classification. Different from the convolutional neural network(CNN) practice where filters are trained by back propagation to represent different patterns of an image, we are inspired by a method called "PCANet" \cite{chan2015pcanet} to choose filter vectors from basis vectors in frequency domain like Fourier coefficients or wavelets without back propagation. Researchers have demonstrated that those basis in frequency domain can usually provide physical insights, which adds to the interpretability of the model by analyzing the frequencies selected. Besides, the training process will also be more time efficient, mathematically clear and interpretable compared with the "black-box" training process of CNN.}

\onecolumn \maketitle \normalsize \setcounter{footnote}{0} \vfill


\section{Introduction}
\label{sec:introduction}

Convolutional Neural Networks(CNN) \cite{jarrett2009best} has witnessed tremendous success in image classification \cite{krizhevsky2012imagenet}, 
with filters performing convolution operations that aim to capture different patterns in an image.
Yet in order to obtain these filter vectors, it is necessary to solve the complicated optimization problem of resorting to back propagation, which makes the whole process a black-box, thereby leading to the lack of clear mathematical interpretations of the resulting filter vectors. 
\cite{chan2015pcanet} put forward a baseline model for image classification, which does not require any kind of back propagation to learn those filters. Instead, they suggested adopting left eigen-vectors of stacked images which are commonly known as principal component vectors as the candidate filters. This idea stems from the eigen-decomposition where we can decompose the target onto the orthogonal basis (eigen-vectors). Projection along each orthogonal basis can represent "non-overlapping" patterns in the image. However, the process of obtaining those eigen-vectors can be quite time-consuming, especially for large datasets, even when some randomized algorithms \cite{halko2011finding} are applied. In the classical literature regarding computer vision, researchers have developed multi-scaled representation of images without resorting to optimization. Two most widely used ones are Discrete Fourier Transformation(DFT) \cite{nordberg1995fourier} and Wavelets analysis \cite{mallat1996wavelets}. \par 

Researchers have found that different frequencies can capture different levels of information in images. For example, the high-pass filter will only select high-frequency signals to get the structured information like edges, while the low-pass filter will select low-frequency signals and thus generate an over-smoothed and blurry image. \cite{costen1996effects}. There are many traditional models focusing on the detection of high-frequency information. For instance, typical gradient-based methods such as the sobel operator \cite{sobel},  prewitt operator \cite{prewitt} and canny operator \cite{canny1986computational} detect the high-frequency information in the 1-order gradient domain. The laplacian operator \cite{wang2007laplacian} focuses on the 2-order gradient, which has been widely applied in image processing to sharpen images. We refer \cite{edgesurvey} for interested readers to gain a comprehensive picture of edge detectors in image processing. In this work, our major focus is the discrete Fourier transformation, since it has a simpler form and can be easily extended to convolutional filters.

In this paper, we shall explore the possibilities of adopting basis from DFT and Wavelets analysis as candidates for filter vectors. Before presenting our algorithms, let us have a brief review of both Discrete Fourier Transformation and Wavelets analysis.

\subsection{Discrete Fourier Transformation}
\label{sec:dft}
Discrete Fourier Transformation (DFT) \cite{beerends2003fourier} can represent vectorized images with different components at different frequencies. Mathematically, given a vectorized image vector $\mathbf{x}$ of length $n$, 
1D DFT decompose $\mathbf{x}$ into Fourier basis $\mathbf{e}^{i\omega_k} =  \mathbf{c}(\omega_k)-i\mathbf{s}(\omega_k)$ with coefficient as the inner product between
 $\mathbf{x}$ and $\mathbf{e}^{i\omega_k}$. Here $\mathbf{s}_k, \mathbf{c}_k$ are defined as
\begin{equation}
\label{eq:cos_sin_coef}
\begin{aligned}
& \mathbf{c}(\omega_k) = \frac{1}{\sqrt{n}} (1, \cos \omega_k, \dots, \cos (n-1)\omega_k)^\top,\\
& \mathbf{s}(\omega_k) = \frac{1}{\sqrt{n}} (1, \sin \omega_k, \dots, \sin (n-1)\omega_k)^\top,
\end{aligned}
\end{equation}
and $\omega_k = \frac{2\pi k}{n}$ are the discrete Fourier frequencies whose index $k$ belongs to a set denoted as $F_n$:  $\left\{-[\frac{n-1}{2}], \dots, [\frac{n}{2}]\right\}$ where $[x]$ is the integer part of $x$. Noticing $\{s_k, c_k, k\in F_n ~\text{and}~ k\ge 0\}$ forms a complete orthogonal basis for $\mathbb{R}^n$ ($\mathbf{s}_0 = \mathbf{0}$), sometimes researchers only consider the non-negative frequencies in the $F_n$. In the following parts, the set containing all non-negative indices in $F_n$ is referred to as $F_n^+$.

\subsection{Wavelets Analysis}
\begin{figure}
  \centering
    \includegraphics[width=0.5\textwidth]{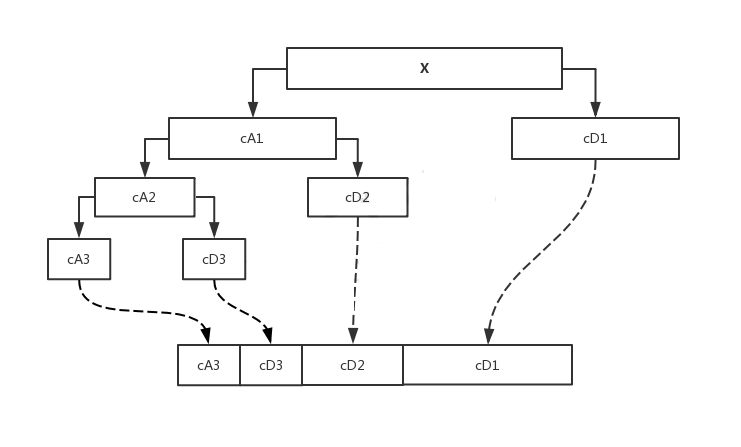}
     \caption{The level-3 wavelet transformation: the first half of the wavelet transformation ($\mathbf{cA_1}$ and $\mathbf{cA_2}$) will be taken to perform the next level transformation, and then concatenate the final level transformation ($\mathbf{cA_3}$ and $\mathbf{cD_3}$) results with the latter half of transformation in the previous levels ($\mathbf{cD_1}$ and $\mathbf{cD_2}$) as the level-3 wavelet transformation result.}
     \label{fig:level-3-wavelet-transformation}
\end{figure}

Different from DFT, wavelets aim to conduct spectral analysis locally in the graph which can be seen from the difference in their orthogonal basis: each Fourier coefficient vector share while wavelets basis vector, which will be introduced later, behaves more abruptly \cite{strang1993wavelet}, \cite{chui2016introduction},  \cite{daubechies1992ten}, \cite{mallat1988multiresolution}. This means wavelets can captures edge information in the computer vision compared to DFT, which will be demonstrated in recover image in Figure \ref{fig:mnist-low-rank-recover}. In this paper we chosee to apply Daubechies D4 Wavelet Transform to the image. 
The wavelet and scaling function coefficients of the Daubechies D4 wavelet are
\begin{equation}
\begin{aligned}
& \mathbf{h} = \left[\frac{1+\sqrt{3}}{4}, \frac{3+\sqrt{3}}{4}, \frac{3-\sqrt{3}}{4}, \frac{1-\sqrt{3}}{4} \right]  \\
& \mathbf{g} = \left[\frac{1-\sqrt{3}}{4} , \frac{\sqrt{3}-3}{4}, \frac{3+\sqrt{3}}{4}, \frac{-1-\sqrt{3}}{4}\right]. 
\end{aligned}
\end{equation}
where $\mathbf{h}$ is the scaling function coefficients and $\mathbf{g}$ is the wavelet function coefficients. $\mathbf{h}$ is like calculating the moving average, which performs as low pass filtering, while $\mathbf{g}$ is capturing the comparison of local graph performing as high pass filtering in the above section. Then for vectorized image $\mathbf{x}$, the first layer wavelet transform is like linear transformation in for n filter vectors as 
\begin{equation}
\label{eq:wave-filter-scale}
\begin{bmatrix}
h_0 & h_1 & h_2 & h_3 & \cdots & \cdots & \cdots \\
0 & 0 & h_0 & h_1 & h_2 & h_3 & \cdots \\
\vdots & \vdots & \vdots & \vdots & \vdots & \vdots & \ddots \\
\end{bmatrix} \mathbf{x}.
\end{equation}
\begin{equation}
\label{eq:wave-filter-wavelet}
\begin{bmatrix}
g_0 & g_1 & g_2 & g_3 & \cdots & \cdots & \cdots\\
0 & 0 & g_0 & g_1 & g_2 & g_3 & \cdots \\
\vdots & \vdots & \vdots & \vdots & \vdots & \vdots & \ddots \\
\end{bmatrix} \mathbf{x}.
\end{equation}
where the first half of the transform is computed via \eqref{eq:wave-filter-scale} and the second half of the transform via \eqref{eq:wave-filter-wavelet}. The second layer wavelet transform is then computed by taking the first half of the first layer transform results, which is results of \eqref{eq:wave-filter-scale}, treat it as an averaged version of the original vectorized image $\mathbf{x}$, and perform linear transformation using row vectors in \eqref{eq:wave-filter-scale} and \eqref{eq:wave-filter-wavelet} with length $\frac{n}{2}$. Then each successive iteration simply repeat the same process. Notice the vector $\mathbf{x}$ and the averaged $\mathbf{x}$ in each layer will be padded to even length if it has an odd length. The final level transformation results and the second half results from the previous levels, which are the results of \eqref{eq:wave-filter-wavelet} of the previous levels, will be concatenated to form the wavelets transformation results. The structure for level-3 wavelet transform is shown in \ref{fig:level-3-wavelet-transformation}.
In this case, we can treat each line of left matrix in \eqref{eq:wave-filter-scale} and \eqref{eq:wave-filter-wavelet} as the pool of our potential filter vectors.

\section{FrequentNet}
Our method is inspired by \cite{chan2015pcanet}, with the difference of adopting basis in frequency domain rather than principal component vectors. 
The overall pipeline is the same as PCANet, which are composed of two procedures. The first procedure is selecting filter vectors, performing convolutional/filter transformation and repeating it once again. (Implementing this procedure once again and presenting its results have also been taken into account). The second procedure is applying hashing and histogram and running the support vector machine to output from the previous procedure and achieve classification.

\subsection{Problem Setup}
We mainly follow the settings in \cite{chan2015pcanet}. Provided with $N$ input training images: $\{\mathcal{I}_i\}_{i=1}^N$
of size $m\times n$, we set the patch size
(or 2D filter size) as $k_1\times k_2$. 
Throughout the paper, padding size has been set to be $(k_1-1)/2$ for the top and the bottom, $(k_2-1)/2$ for the left and the right, with the padding value at zero. Meanwhile, stride for patch is set to be one, which is also listed in Table \ref{tab:one-stage-setup}. Under this setting after the filter transformation, the output size will be the same as the input image size: $m\times n$. We call those vectorized patches $\mathbf{x}_{i, 1}, \cdots, \mathbf{x}_{i, mn}$ where the first index is for images and the second index is for patches. The patch mean has been subtracted from each patch and stacked for each image as
\begin{equation}
\mathbf{\bar{X}}_i = [\mathbf{\bar{x}}_{i,1}, \cdots, \mathbf{\bar{x}}_{i,j}, \cdots, \mathbf{\bar{x}}_{i,mn}], 1\le j\le mn
\end{equation}
where $\mathbf{\bar{x}}_{i,j}$ is the de-meaned patch. We further stack them as 
$\mathbf{\bar{X}}_i$ again to obtain
\begin{equation}
\bar{\mathbf{X}} = [\bar{\mathbf{X}}_1, \cdots, \mathbf{\bar{X}}_i, \cdots, \mathbf{\bar{X}}_N] \in \mathbb{R}^{k_1k_2\times Nmn}, 1\le i\le N.
\end{equation}
Filter vectors are intended for representing patterns in columns in $\mathbf{x}_{i,j}$ effectively. PCANet chooses the filter vectors to be the top left eigen-vectors of $\mathbf{\bar{X}}$. In this paper, adopting basis in DFT and wavelets have been proposed in a detailed manner.

\subsection{FourierNet}
\paragraph{The First Stage:} We choose $\{\mathbf{c}(\omega_k), \mathbf{s}(\omega_k)\}, k\in F_n^+$ as our candidate orthogonal basis, and then select filters at different frequencies based on the magnitude of the inner product of vectorized patches $\mathbf{x}_{i,j}$ and candidate filters, as summarized in Algorithm \ref{alg:f_topk}. With the obtained $L_1$ filters $\mathbf{v}_1, \cdots, \mathbf{v}_k, \cdots, \mathbf{v}_{L_1}$, every input image $\mathcal{I}_i$ is mapped to $L_1$ new feature maps:
\begin{equation}
\mathcal{I}^\ell_i = \mathcal{I}_i * \rm{mat}_{k_1, k_2}(\mathbf{v_\ell}),  
\end{equation}
where $*$ is the two dimensional convolution and mat is an operator reshaping filter back to its original shape of the patch: $k_1\times k_2$. For convenience, we index the vector set $\mathbf{v} \in \mathcal{D}_{L_1}$ based on $\|\langle \mathbf{v}, \mathbf{\bar{X}}\rangle \|$ reversely, i.e., 
$\|\langle \mathbf{v}_1, \mathbf{\bar{X}}\rangle \| \ge \cdots \ge  \|\langle \mathbf{v}_{L_1}, \mathbf{\bar{X}} \rangle\|$.

\begin{algorithm}[tb]
   \caption{Select top K Fourier Basis}
   \label{alg:f_topk}
\begin{algorithmic}
   \STATE {\bfseries Input:} $\mathbf{\bar{X}}, L_1$
   \FOR{$k$ in $F_{k_1k_2}^{+}$}
   \STATE  $c_k \leftarrow  \| \langle \mathbf{c}(\omega_k), \mathbf{\bar{X}} \rangle\|$
   \STATE  $s_k\leftarrow \|\langle \mathbf{s}(\omega_k), \mathbf{\bar{X}} \rangle \|$
   \ENDFOR
   \STATE Select $\mathbf{c}(\omega_k) ~\rm{or}~ \mathbf{s}(\omega_k)$ with top $L_1$ largest corresponding values in $c_k, s_k$ 
   and call the set of $\mathcal{D}_{L_1}$
   \STATE {\bfseries Output:} $\mathcal{D}_{L_1}$
\end{algorithmic}
\end{algorithm}

\paragraph{The Second Stage:} After the first stage, for each filter vector $\mathbf{v}_\ell$ in $\mathcal{D}_{L_1}$, we can get a new set of feature matrices of the same size as the original image: $m\times n$. For the new $L_1N$ feature matrices denoted by $\mathcal{I}^\ell_i \in \mathbb{R}^{m\times n}, i=1, \cdots, N, \ell=1,\cdots L_1$, steps in the first stage are repeated to continue to stack all the overlapping patches and subtract mean from them. For each filter $\mathbf{v}_\ell, 1\le \ell \le L_1$, by stacking Nmn de-meaned patches we can get $\mathbf{\bar{Y}}^\ell$, like $\mathbf{\bar{X}}$ defined in the first stage:
\begin{equation}
\mathbf{\bar{Y}}^\ell = [\mathbf{\bar{y}}^\ell_{i,1}; \cdots; \mathbf{\bar{y}}^\ell_{i,mn}] \in \mathbb{R}^{k_1k_2\times Nmn}, 1\le i\le N
\end{equation}
where $\mathbf{\bar{y}}^\ell_{i,j}$ is the de-meaned patch in the $\mathcal{I}_i$. Then we stack all $\mathbf{\bar{Y}}^\ell$ to obtain our final feature matrix 
\begin{equation}
\mathbf{\bar{Y}} = [\mathbf{\bar{Y}}^1; \cdots; \mathbf{\bar{Y}}^\ell \cdots; \mathbf{\bar{Y}}^{L1}]\in \mathbb{R}^{k_1k_2\times L_1Nmn}.
\end{equation}
Algorithm \ref{alg:f_topk} is applied on $\mathbf{\bar{Y}}$ to select top $L_2$ Fourier basis. Following the above definition, we call selected basis $\mathbf{u}_1, \cdots, \mathbf{u}_{L_2}$. According to our setting in stride and padding, the output feature matrix is still of the same size as the original input image: $m\times n$. We call it the output feature matrix $\mathcal{O}_i^\ell$:
\begin{equation}
\mathcal{O}_i^{\ell_2} = \mathcal{I}_i^{\ell_1} * \rm{mat}_{k_1, k_2}(\mathbf{u}_{\ell_2}), 1\le \ell_1\le L_1, 1\le \ell_2 \le L_2.
\end{equation}

\paragraph{Output Stage:}
In this section, we follow exactly the procedure of output stage in \cite{chan2015pcanet} to finally transform the output of feature matrix, and run support vector machine to make the prediction. Steps have been briefly sketched. For more details, we refer readers to \cite{chan2015pcanet}. This procedure works for both the first and second stages, with notations used in the two stage model for explanation. Given the $NL_1$ input feature 
matrix: $I_i^{\ell_1}$ and selected $L_2$ filters $\mathbf{u}_{\ell_2}$, we aggregate filter information as 
\begin{equation}
\mathcal{T}_i^{\ell_1} = \Sigma_{\ell_2=1}^{L_2} 2^{\ell_2-1}H(\mathcal{I}_i^{\ell_1} * \mathbf{u}_{\ell_2})
\end{equation}
where H is binary operator turning positive elements to be one while others to zero element wise. It is easy to see each element fall into $[0, 2^{L_2}-1]$. \par 
For each patch in $\mathcal{T}_i^{\ell_1}$, we computer the histogram ($2^{L_2}$ bins) with two parameters called block size and block stride. Concatenating all histograms for each $\mathcal{T}_i^{\ell_1}$ of $\rm{Bhist}(\mathcal{T}_i^{\ell_1})$ across $L_1$ filters, we can obtain the feature vector 
\begin{equation}
\mathbf{f}_i = \left[\rm{Bhist}(\mathcal{T}_i^1), \cdots, \rm{Bhist}(\mathcal{T}_i^{L_1})\right] \in \mathbb{R}^{2^{(2^{L_2})}L_1B}. 
\end{equation}
Then we run support vector machine on feature vector $\mathbf{f}_i$ to obtain the classification results.

\subsection{WaveletsNet}
For WaveletsNet, the whole process is the same as FourierNet except that now the pool of candidate filter vectors become all rows in orthogonal basis in wavelets.
In this paper, we only consider basis from three layers in Daubechies D4 wavelets. Again, for stage I, $L_1$ filter vectors are selected based on the magnitude of inner product between filter vectors and vectorized image vectors. At this point, we can either go to the output stage, or repeat the selecting procedure to further select $L_2$ filters and then move on to the output stage.

\subsection{PCANet, RandNet}
In \cite{chan2015pcanet}, researchers proposed PCANet where the filter candidates are from PCA vectors for $\mathbf{\bar{X}}$ in the first stage and for $\mathbf{\bar{Y}}$ in the second stage. With the importance of PCA vectors naturally sorted by corresponding eigen-values, we just choose the top $L_1$ or $L_2$ PCA vectors. For RandNet, as the name suggests, filter vectors are chosen randomly from multivariate Gaussian distributions. By comparing our method with those two methods, it is worth noting that unlike PCANet, filters in FreqNet are data independent. 


\section{Experiments}
We evaluated the performances of proposed models, which are described in the Table \ref{tab:structure-desciprtion}, and compared with PCANet and RandNet on two tasks, the hand-written digits recognition and object recognition. Then we analyzed the extracted filters in FourierNet and WaveletsNet by visualizing the filtered image by selected filters. We use level-1 Daubechies D4 wavelet transformation for all kinds of WaveletsNet below during the experiments. The code for the model and experiments are publicly available \footnote[1]{\url{https://github.com/ijcaiworkshop2020/freqnet}}.

\begin{table}
\caption{Model Structure Description}
\label{tab:structure-desciprtion}
\small
\centering
\begin{tabular}{@{}ll@{}}
\toprule
Model   & Description                              \\
\midrule
FourierNet-1      & 1-stage FourierNet       \\
FourierNet-2    & 2-stage FourierNet              \\
WaveletsNet-1        & 1-stage FourierNet                      \\
WaveletsNet-2    & 2-stage FourierNet              \\
PCANet-1        & 1-stage PCANet \\
PCANet-2        & 2-stage PCANet \\
RandNet-1       & 1-stage RandNet \\
RandNet-2       & 2-stage RandNet \\
FourierNet2D-2  & 2-stage FourierNet (2D Fourier basis) \\ \bottomrule
\end{tabular}
\end{table}

\subsection{Hand-written Digits Recognition}
We first conducted the experiment on MNIST variations. The MNIST \cite{lecun1998gradient} and MNIST variations \cite{larochelle2007empirical} are common benchmarks for testing hierarchical representations \cite{chan2015pcanet}. We pick a subset of MNIST variations to experiment on. The datasets and their descriptions are listed in the Table \ref{tab:mnist-desciprtion}.

\begin{table}
\centering
\caption{Descriptions of MNIST variations picked for experiment}
\label{tab:mnist-desciprtion}
\small
\begin{tabular}{@{}ll@{}}
\toprule
Dataset   & Description                              \\ \midrule
basic      & A smaller subset of standard MNIST       \\
bg-rand    & MNIST with noise background              \\
rot        & MNIST with rotation                      \\
bg-img     & MNIST with image background              \\
bg-img-rot & MNIST with rotation and image background \\ \bottomrule
\end{tabular}
\end{table}

\subsubsection{Experiment Setup}
Following the experiment setup in \cite{chan2015pcanet}, we conducted experiments using both one-stage and two-stage models. For one-stage model, we fixed patch size to $7 \times 7$, then investigated the impact of number of filters $L_1$. For two-stage models, we followed the recommended configurations for different datasets in the original PCANet paper. Notice instead of using the parameter block overlap ratio in the block-wise histogram stage, we explicitly define the block stride by computing the overlap and truncated to the nearest integer. For all the experiments, we fixed patch stride to $1$ and padded zero around the image in the patch collection stage. Other detailed experiment parameters, for both one-stage and two-stage models, are listed in Table \ref{tab:one-stage-setup} and Table \ref{tab:two-stage-setup} respectively.


\begin{table}
\centering
\caption{Experiment setup of one-stage models}
\label{tab:one-stage-setup}
\begin{tabular}{@{}llllll@{}}
\toprule
Dataset  & $L_1$ & patch size  & \shortstack{block\\size} & \shortstack{block\\stride} \\ \midrule
basic     & 8     & 7$\times$7 & 7$\times$7        & 3            \\
bg-rand   & 8     & 7$\times$7 & 4$\times$4        & 2            \\
rot       & 8     & 7$\times$7 & 4$\times$4        & 2            \\
bg-img    & 8     & 7$\times$7 & 4$\times$4        & 2            \\
bg-img-rot & 8    & 7$\times$7 & 4$\times$4        & 2            \\ \bottomrule
\end{tabular}
\end{table}

\begin{table}
\centering
\caption{Experiment setup of two-stage models}
\label{tab:two-stage-setup}
\begin{tabular}{@{}lllllll@{}}
\toprule
Dataset   & $L_1$ & \shortstack{patch\\size} & $L_2$  & \shortstack{block\\size} & \shortstack{block\\stride} \\ 
\midrule 
basic      & 6 & 8 & 7$\times$7 & 7$\times$7 & 3 \\
bg-rot     & 6 & 8 & 7$\times$7 & 4$\times$4 & 2 \\
bg-rand    & 6 & 8 & 7$\times$7 & 4$\times$4 & 2 \\
bg-img     & 6 & 8 & 7$\times$7 & 4$\times$4 & 2 \\
bg-img-rot & 6 & 8 & 7$\times$7 & 4$\times$4 & 2 \\ \bottomrule
\end{tabular}
\end{table}

\subsubsection{Experiment Results}
\begin{figure}
    \centering
    \includegraphics[width=0.51\textwidth]{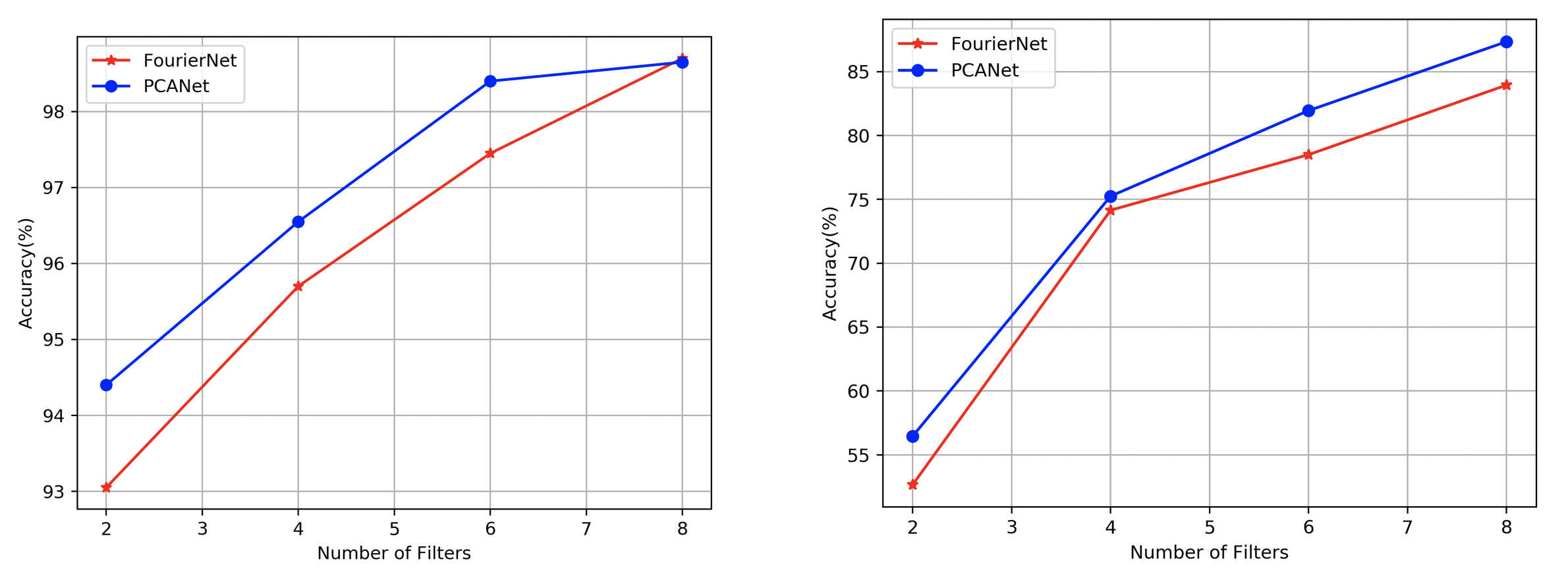}
    \caption{Test accuracy(\%) of FourierNet-1 and PCANet-1 on MNIST basic and rot test set for varying number of filters($L_1$). We tested $L_1$ varies from 2 to 8.}
  \label{fig:one-stage-accuracy-vs-l1}
\end{figure}

The test accuracy of the one stage models on the selected datasets, with the number of filters varies from 2 to 8 are shown in Figure \ref{fig:one-stage-accuracy-vs-l1}. We can see from the results that the test accuracy increases when the number of filters grows. 
The test results for the two-stage models, with the key setup in Table \ref{tab:two-stage-setup} are listed in Table \ref{tab:two-stage-mnist-accuracy}. One can see that FourierNet-2 and WaveletsNet-2 achieve similar testing accuracy on these datasets.

\begin{table}
\centering
\caption{Testing Accuracies(\%) of different models of MNIST variations. The model parameters follow Table \ref{tab:one-stage-setup} and Table \ref{tab:two-stage-setup}}
\label{tab:two-stage-mnist-accuracy}
\small
\begin{tabular}{@{}llllll@{}}
\toprule
Model      & basic & \shortstack{bg-\\rand} & rot   & \shortstack{bg-\\img} & \shortstack{bg-\\img-rot} \\ \midrule
FourierNet-2 & 98.05 & 90.50    & 89.45 & 86.55  & 60.25      \\
FourierNet-1 & 98.75  & 89.95   & 85.15 & 85.45  & 49.65      \\
WaveletsNet-2    & 98.55 & 88.05    & 83.60  & 84.45  & 49.25      \\
WaveletsNet-1    & 97.55  & 83.60   & 83.60  & 78.05  & 42.20      \\
PCANet-2     & 98.15  & 91.55    & 89.50  & 87.20   & 62.50       \\
PCANet-1     & 98.65 & 91.80    & 87.35 & 86.65  & 54.35      \\
RandNet-2    & 97.55 & 82.70    & 86.00  & 83.40   & 40.00       \\
RandNet-1    & 97.95 & 70.90    & 77.55 & 67.55  & 29.90       \\ \bottomrule
\end{tabular}
\end{table}

\subsubsection{Discussion}
\begin{figure}
  \centering
    \includegraphics[width=0.5\textwidth]{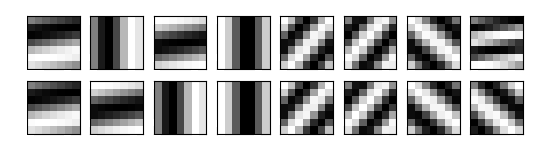}
      \caption{The Fourier filters learned from \emph{bg-rand} dataset. Top: the first stage filters. Bottom: the second stage filters.}
  \label{fig:bg-rand-Fourier-filter}
\end{figure}

To understand what the proposed models are learning, we listed the learned first and second stage Fourier filters from \emph{bg-rand} dataset in Figure \ref{fig:bg-rand-Fourier-filter}. We could see from the visualized filters that most of them are on the low frequency side, like the first four columns of Figure \ref{fig:bg-rand-Fourier-filter}. Furthermore, we investigated the selected frequencies for other MNIST variations, and we observe that the low frequencies , such as $2\pi /49$, $12\pi / 49$ and $14\pi / 49$ appear in selected frequencies for all the datasets. Hence, we believe that the low frequency components carry significant information for the MNIST variations. We report the selected Fourier basis in FourierNet-2 for MNIST variations in Table \ref{tab:selected-fourier-basis}.

\begin{figure}
    \centering 
\begin{subfigure}{0.10\textwidth}
  \includegraphics[width=\linewidth]{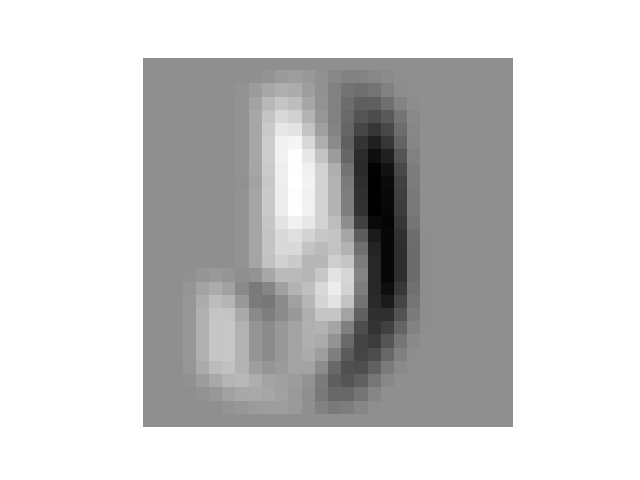}
  \includegraphics[width=\linewidth]{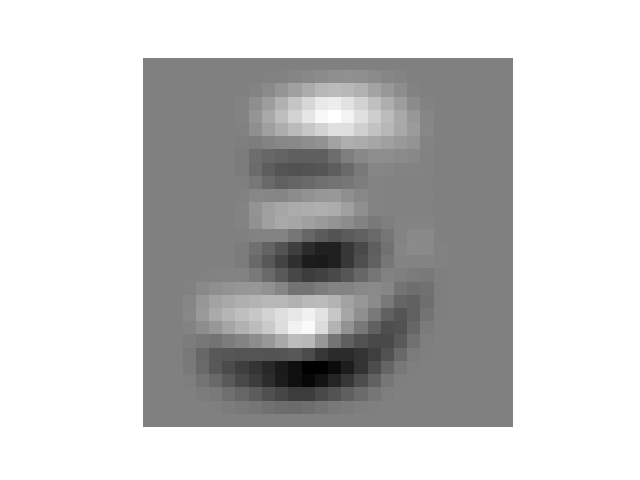}
  \includegraphics[width=\linewidth]{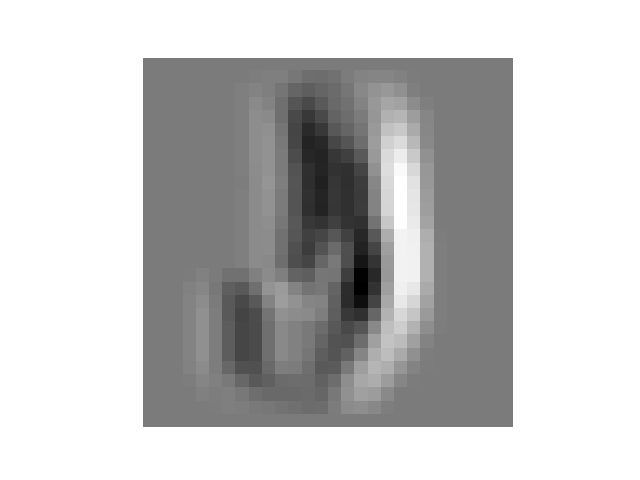}
  \includegraphics[width=\linewidth]{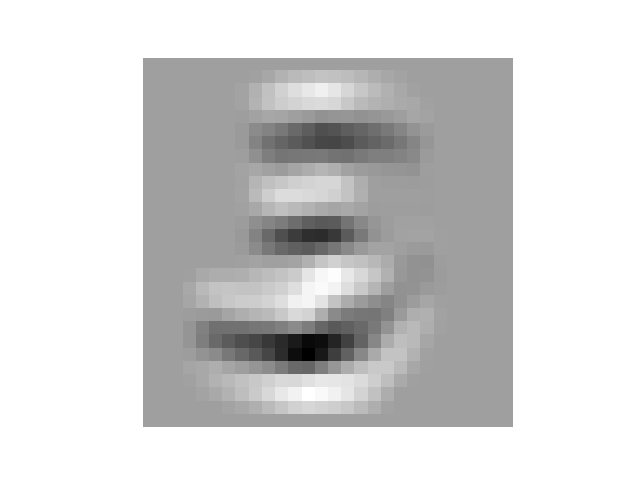}
  \includegraphics[width=\linewidth]{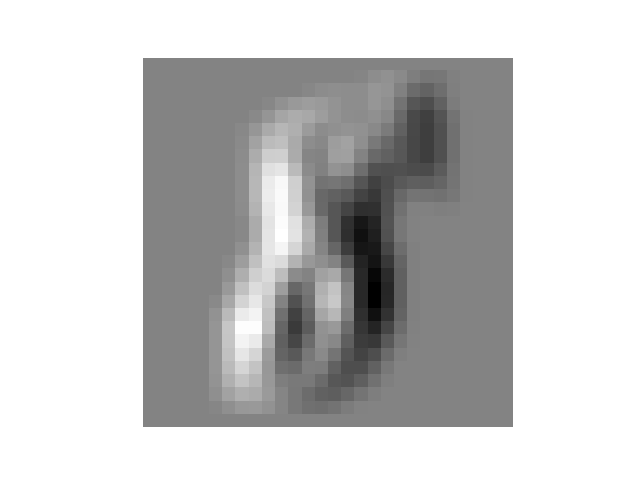}
  \includegraphics[width=\linewidth]{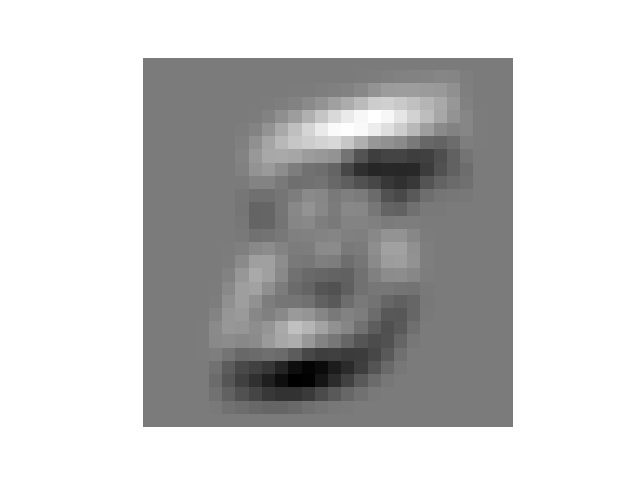}
  \includegraphics[width=\linewidth]{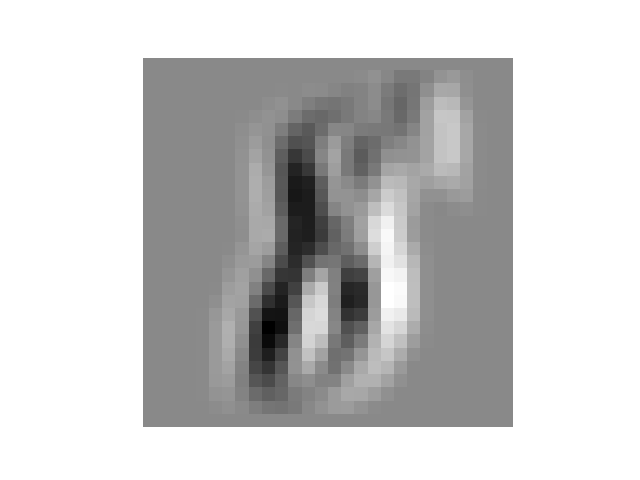}
  \includegraphics[width=\linewidth]{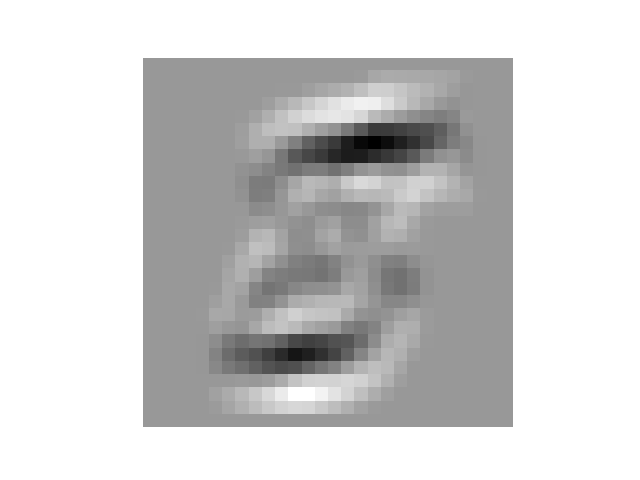}
  \caption{Fourier}
  \label{fig:low-rank-recover-Fourier}
\end{subfigure}\hfil 
\begin{subfigure}{0.10\textwidth}
  \includegraphics[width=\linewidth]{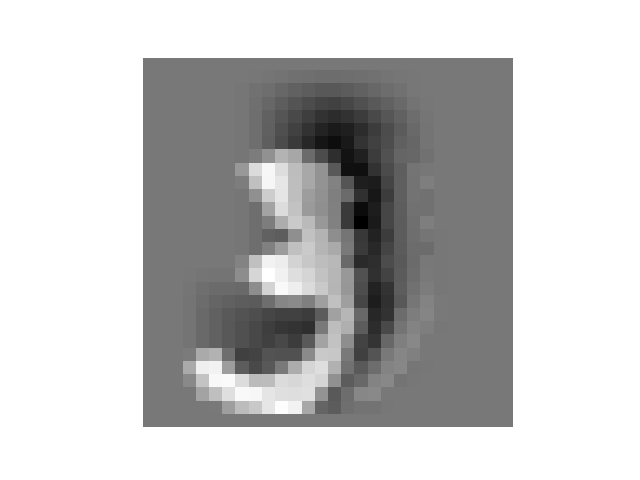}
  \includegraphics[width=\linewidth]{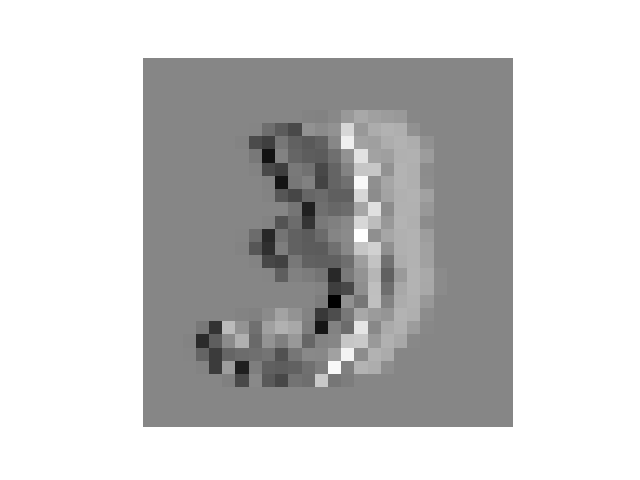}
  \includegraphics[width=\linewidth]{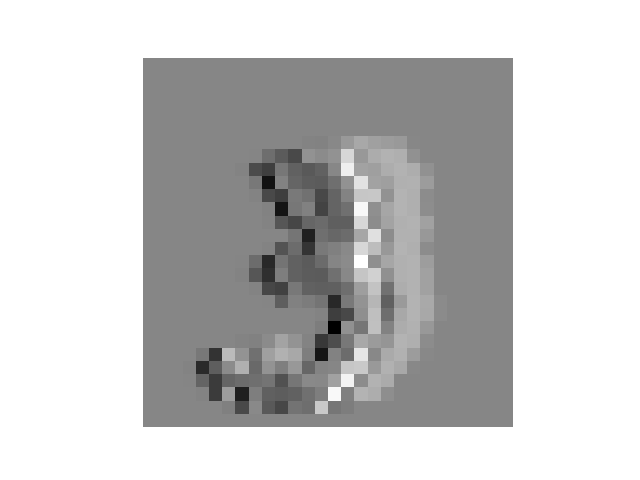}
  \includegraphics[width=\linewidth]{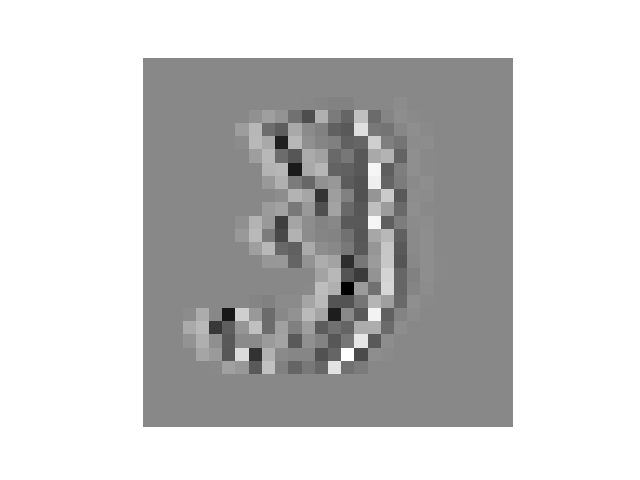}
  \includegraphics[width=\linewidth]{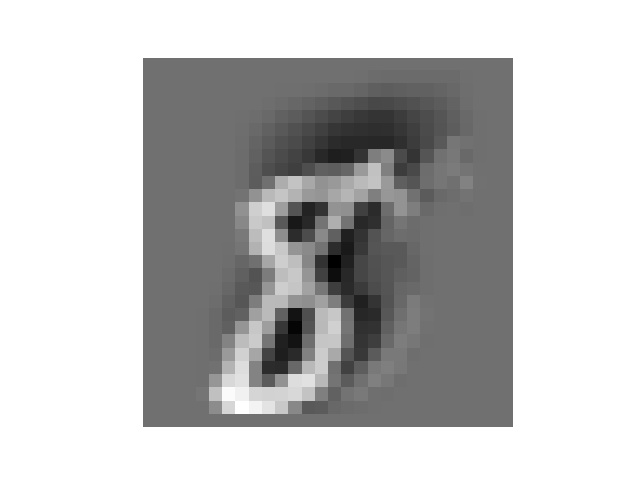}
  \includegraphics[width=\linewidth]{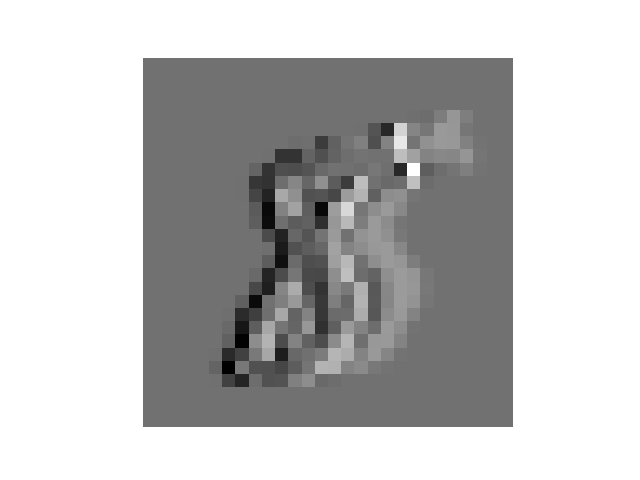}
  \includegraphics[width=\linewidth]{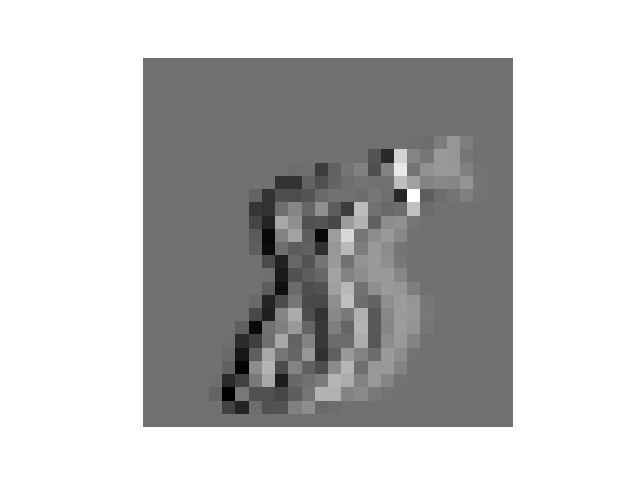}
  \includegraphics[width=\linewidth]{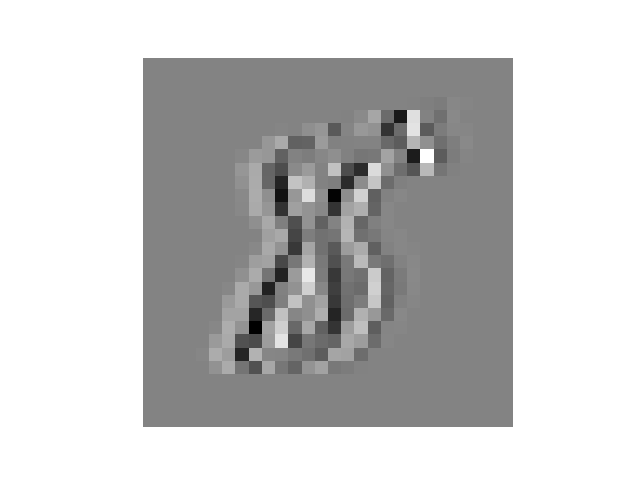}
  \caption{Wavelet}
  \label{fig:low-rank-recover-wavelet}
\end{subfigure}\hfil 
\begin{subfigure}{0.10\textwidth}
  \includegraphics[width=\linewidth]{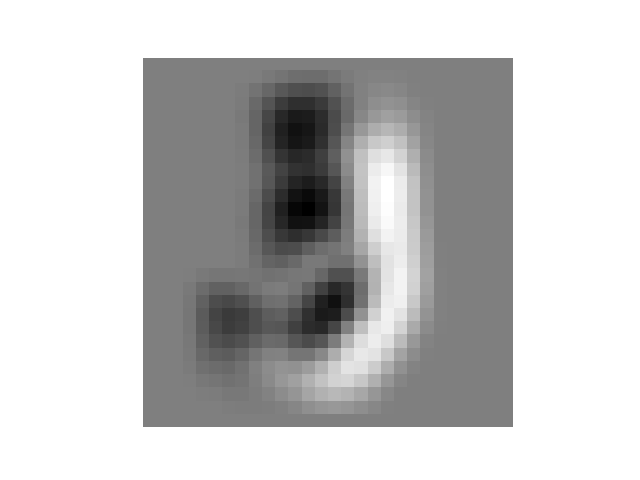}
  \includegraphics[width=\linewidth]{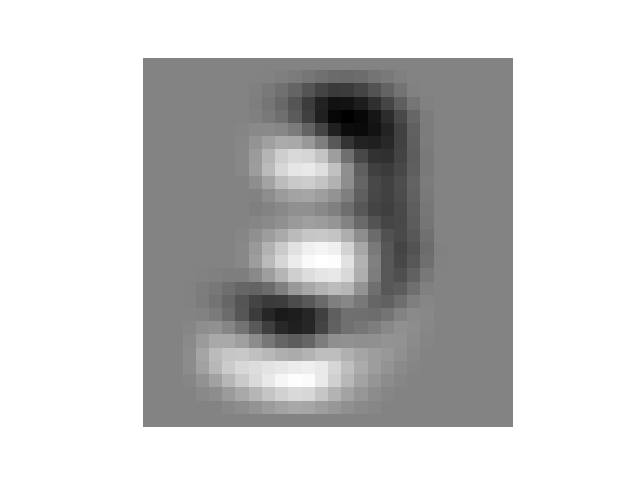}
  \includegraphics[width=\linewidth]{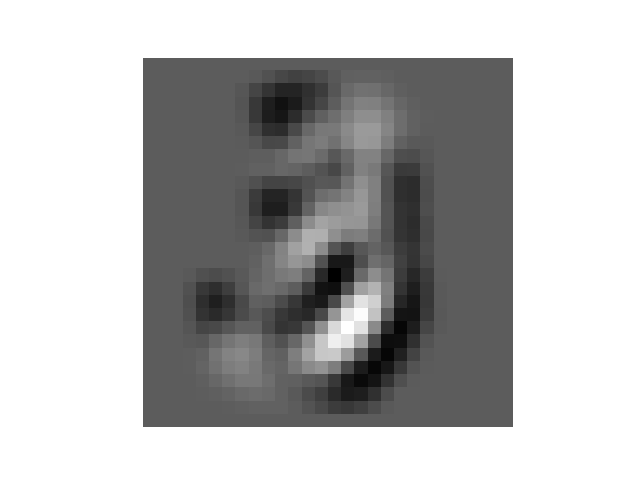}
  \includegraphics[width=\linewidth]{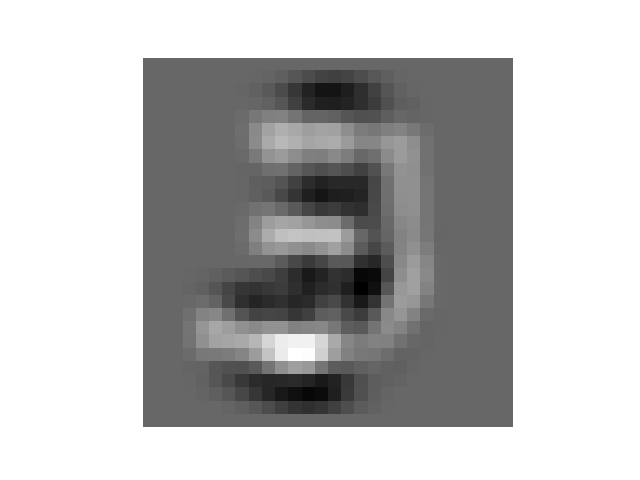}
  \includegraphics[width=\linewidth]{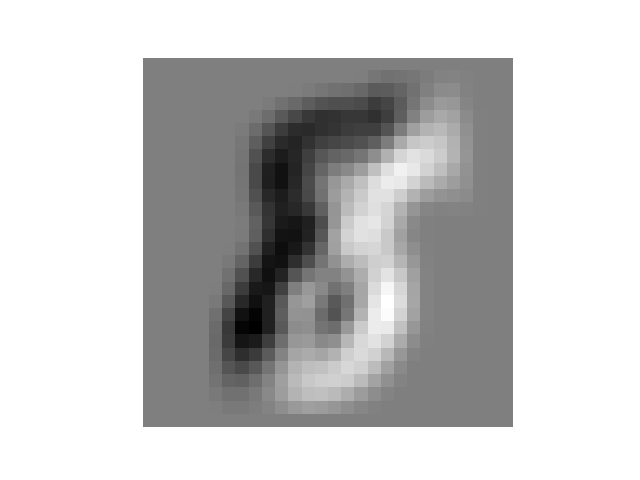}
  \includegraphics[width=\linewidth]{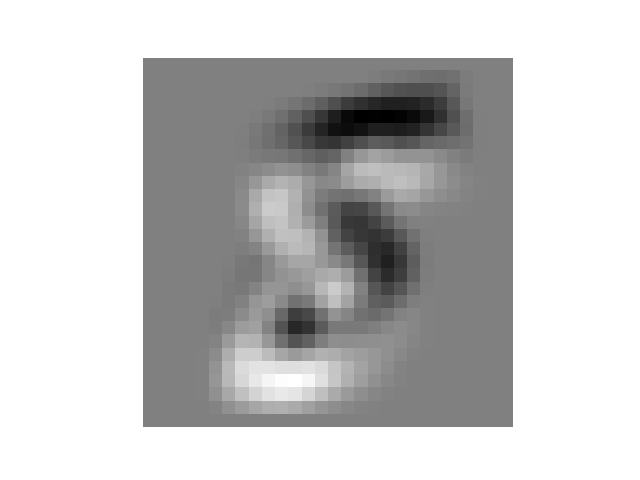}
  \includegraphics[width=\linewidth]{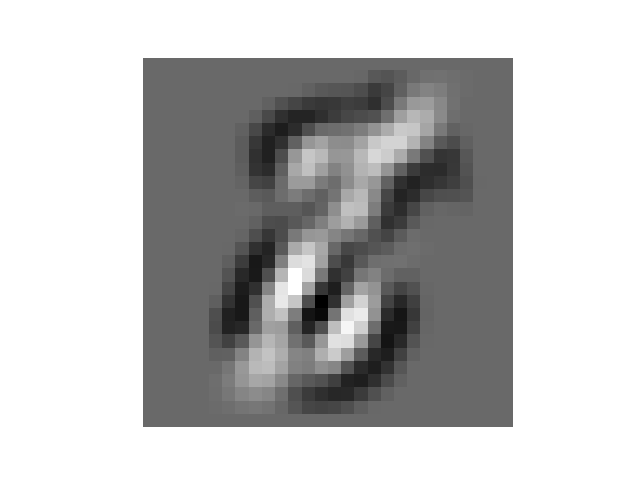}
  \includegraphics[width=\linewidth]{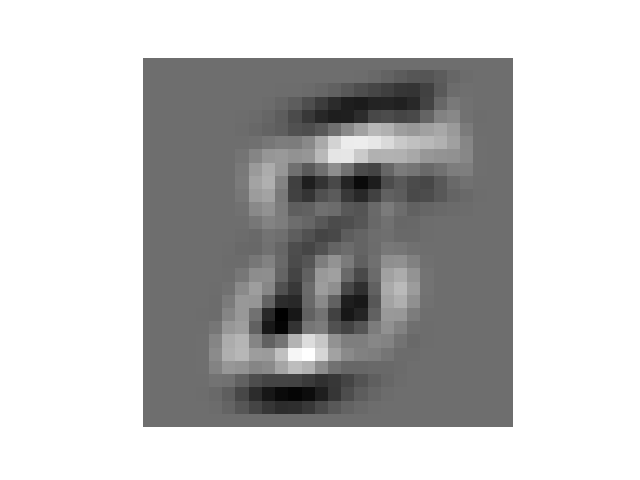}
  \caption{PCA}
  \label{fig:low-rank-recover-PCA}
\end{subfigure}\hfil 
\begin{subfigure}{0.10\textwidth}
  \includegraphics[width=\linewidth]{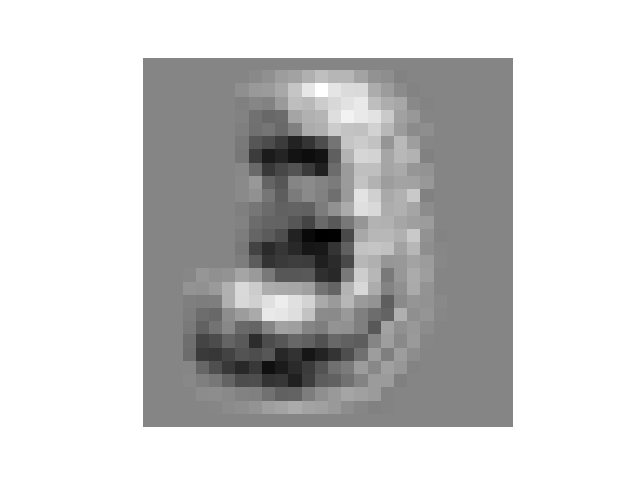}
  \includegraphics[width=\linewidth]{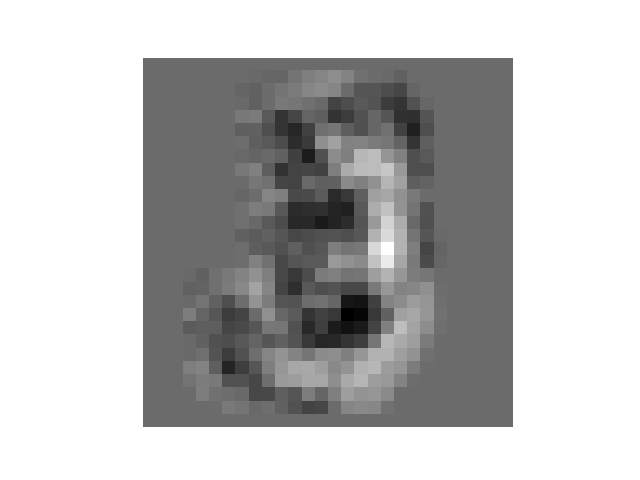}
  \includegraphics[width=\linewidth]{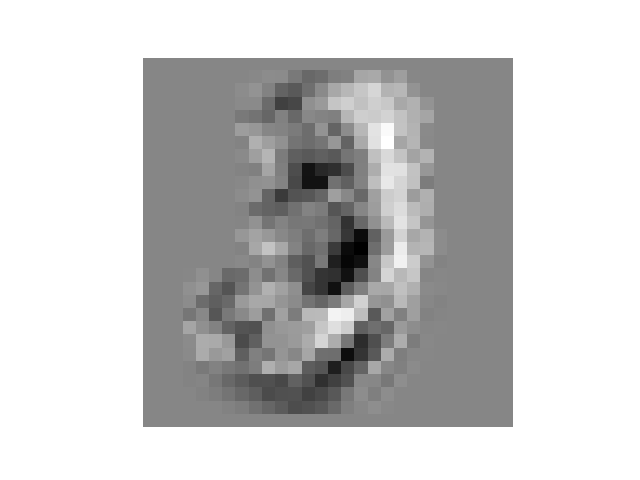}
  \includegraphics[width=\linewidth]{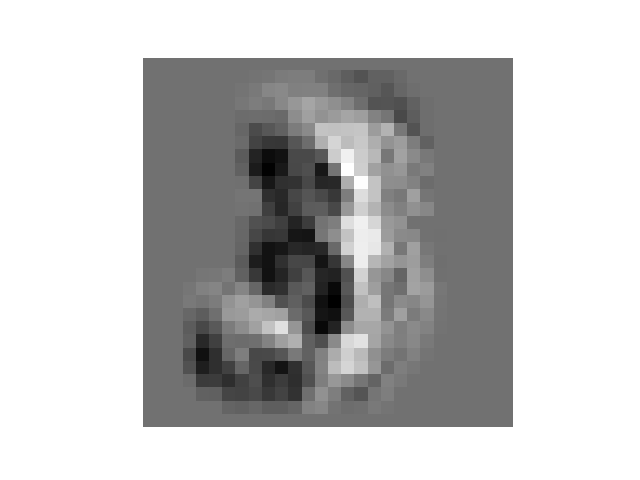}
  \includegraphics[width=\linewidth]{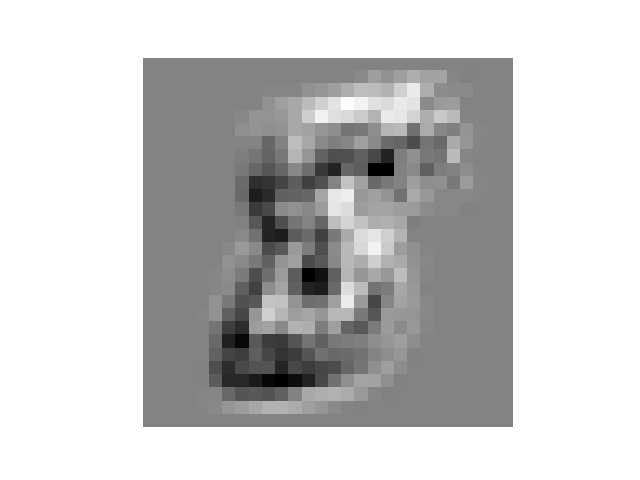}
  \includegraphics[width=\linewidth]{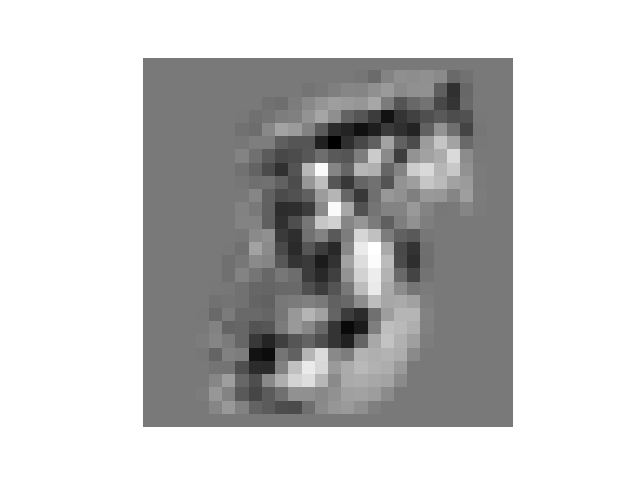}
  \includegraphics[width=\linewidth]{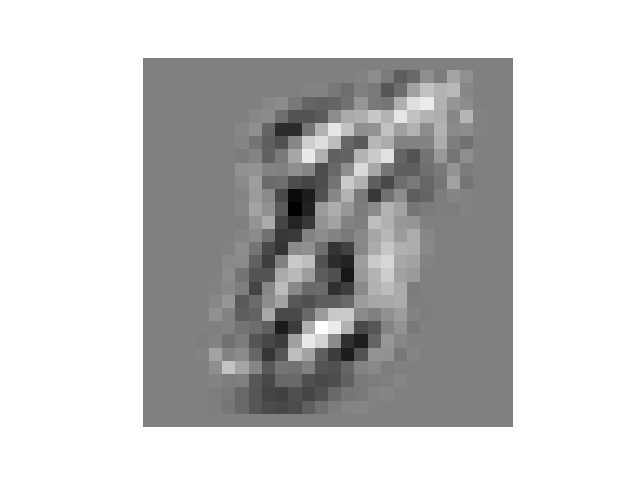}
  \includegraphics[width=\linewidth]{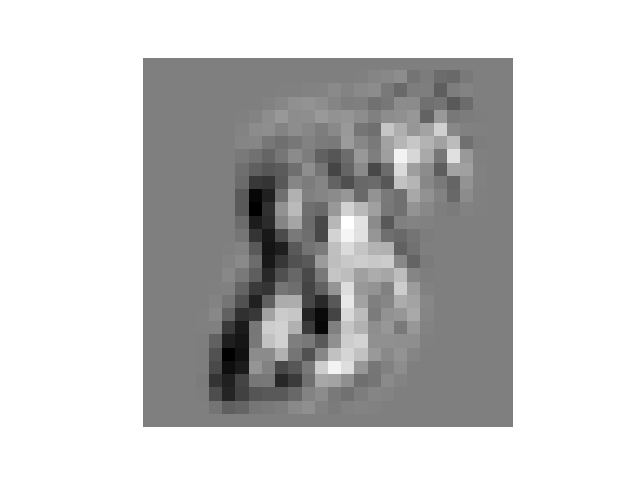}
  \caption{Random}
  \label{fig:low-rank-recover-random}
\end{subfigure}\hfil 
\caption{The low rank approximations using a single filter learned from the network, from left to right: (a) Fourier filter; (b) Wavelet filter (c) PCA filter; (d) Random filter. We picked four selected filters for each basis. Then, for each of the selected filters, we performed convolution with the two randomly selected $28 \times 28$ image samples. From the above figures, one could see that Fourier filters and PCA filters extract features with similar patterns, and selected wavelet filters extract "edge-like" features from the images.}
\label{fig:mnist-low-rank-recover}
\end{figure}

In order to get an intuitive sense of the features captured by the learned filters, we visualized some of the convolution results of an original image from dataset and a single filter selected by the models. Two samples are selected randomly from MNIST dataset, and are convoluted with learned filters, including Fourier filter, Wavelet filter, PCA filter and Random filter. The feature captured by the learned filters are shown in Figure \ref{fig:mnist-low-rank-recover}, each of the images is the convolution results of a raw image sample with a selected filter.

\begin{table*}[t]
    \centering
    \scriptsize
    \begin{tabular}{c|cccccc|cccccccc}
    \toprule
    Dataset     & \multicolumn{6}{c}{Selected first stage basis} 
    & \multicolumn{8}{|c}{Selected second stage basis} \\ 
    \midrule
    basic       & $\mathbf{c}(\omega_1)$ & $\mathbf{c}(\omega_6)$ & $\mathbf{c}(\omega_7)$  & $\mathbf{s}(\omega_1)$ & $\mathbf{s}(\omega_6)$ & $\mathbf{s}(\omega_7)$  & $\mathbf{c}(\omega_1)$ & $\mathbf{c}(\omega_6)$ & $\mathbf{c}(\omega_7)$  & $\mathbf{c}(\omega_8)$ & $\mathbf{s}(\omega_1)$ & $\mathbf{s}(\omega_6)$ & $\mathbf{s}(\omega_7)$ & $\mathbf{s}(\omega_8)$ \\ \cline{2-15}  
    bg-rand       & $\mathbf{c}(\omega_1)$ & $\mathbf{c}(\omega_6)$ & $\mathbf{c}(\omega_7)$  & $\mathbf{s}(\omega_1)$ & $\mathbf{s}(\omega_6)$ & $\mathbf{s}(\omega_7)$  & $\mathbf{c}(\omega_1)$ & $\mathbf{c}(\omega_6)$ & $\mathbf{c}(\omega_7)$  & $\mathbf{c}(\omega_8)$ & $\mathbf{s}(\omega_1)$ & $\mathbf{s}(\omega_6)$ & $\mathbf{s}(\omega_7)$ & $\mathbf{s}(\omega_8)$ \\ \cline{2-15}  
    rot       & $\mathbf{c}(\omega_1)$ & $\mathbf{c}(\omega_6)$ & $\mathbf{c}(\omega_7)$  & $\mathbf{s}(\omega_1)$ & $\mathbf{s}(\omega_6)$ & $\mathbf{s}(\omega_7)$  & $\mathbf{c}(\omega_1)$ & $\mathbf{c}(\omega_6)$ & $\mathbf{c}(\omega_7)$  & $\mathbf{c}(\omega_8)$ & $\mathbf{s}(\omega_1)$ & $\mathbf{s}(\omega_6)$ & $\mathbf{s}(\omega_7)$ & $\mathbf{s}(\omega_8)$ \\ \cline{2-15}  
    bg-img       & $\mathbf{c}(\omega_1)$ & $\mathbf{c}(\omega_6)$ & $\mathbf{c}(\omega_7)$  & $\mathbf{s}(\omega_1)$ & $\mathbf{s}(\omega_6)$ & $\mathbf{s}(\omega_7)$  & $\mathbf{c}(\omega_1)$ & $\mathbf{c}(\omega_6)$ & $\mathbf{c}(\omega_7)$  & $\mathbf{c}(\omega_8)$ & $\mathbf{s}(\omega_1)$ & $\mathbf{s}(\omega_6)$ & $\mathbf{s}(\omega_7)$ & $\mathbf{s}(\omega_8)$ \\ \cline{2-15}  
    bg-img-rot       & $\mathbf{c}(\omega_1)$ & $\mathbf{c}(\omega_6)$ & $\mathbf{c}(\omega_7)$  & $\mathbf{s}(\omega_1)$ & $\mathbf{s}(\omega_6)$ & $\mathbf{s}(\omega_7)$  & $\mathbf{c}(\omega_1)$ & $\mathbf{c}(\omega_6)$ & $\mathbf{c}(\omega_7)$  & $\mathbf{c}(\omega_9)$ & $\mathbf{s}(\omega_1)$ & $\mathbf{s}(\omega_6)$ & $\mathbf{s}(\omega_7)$ & $\mathbf{s}(\omega_9)$ \\ \hline  
      \end{tabular}
    \caption{Selected Fourier basis in FourierNet-2 for different datasets. Following the definition in Section \ref{sec:dft}, $\mathbf{c}(\omega_k)$ and $\mathbf{s}(\omega_k)$ are the orthogonal Fourier basis, where $\omega_k = \frac{2k\pi}{49}$ since the patch size for all the MNIST variation datasets is $49~(7 \times 7)$. The model parameters follow Table \ref{tab:two-stage-setup}.}
    \label{tab:selected-fourier-basis}
\end{table*}

\subsubsection{Extension}
 We further constructed FourierNet2D, which is a two-stage FourierNet using 2D Fourier basis, and tested on MNIST variations. The results are shown in Table \ref{tab:two-stage-mnist-accuracy-2d-Fourier}. However, we did not observe improvements in test accuracy using 2D Fourier basis.

Some related research can be found in \cite{zhu2021news} \cite{zhu2020high} \cite{zhu2020adaptive} \cite{zhu2021time} \cite{zhu2021clustering} \cite{jarrow2021low} \cite{sun2021tensor} \cite{sun2020low}.

\begin{table}
\centering
\small
\caption{Testing Accuracies(\%) of FourierNet2D-2}
\label{tab:two-stage-mnist-accuracy-2d-Fourier}
\begin{tabular}{@{}llllll@{}}
\toprule
Model      & basic & \shortstack{bg-\\rand} & rot   & \shortstack{bg-\\img} & \shortstack{bg-\\img-rot} \\ \midrule
FourierNet2D-2 & 97.80 & 89.60    & 87.60 & 86.05  & 45.30      \\
 \bottomrule
\end{tabular}
\end{table}

\subsection{CIFAR10 Object Recognition}
\subsubsection{Experiment Results}
CIFAR10 contains 10 classes with 50000 training samples and 10000 test samples, which vary in object position, scale, colors and textures \cite{chan2015pcanet}.
We fix the number of filters in the first stage to 40, the number of filters in second stage to 5. We also set the patch size to 5 $\times$ 5, block size to 8 $\times$ 8 and block stride to 4. Apart from the two-stage FourierNet and two-stage WaveletsNet, we tried combining Fourier basis and Wavelets basis together to form two combined two-stage models, namely Fourier-PCA, which uses Fourier filters in the first stage and PCA filters for the second stage, and PCA-Fourier, which uses PCA filters in the first stage and Fourier filters in the second stage. The test accuracy of different combinations are listed in Table \ref{tab:accuracy-cifar10}.

\subsubsection{Discussion}
\begin{figure}
  \centering
    \includegraphics[width=0.5\textwidth]{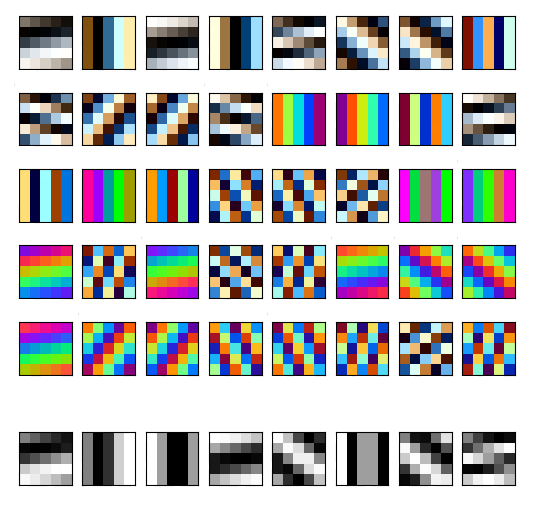}
     \caption{The Fourier filters learned from \emph{CIFAR10} dataset. Top: the first stage filters, the number of filters for each channel is set to 40. Bottom: the second stage filters, the number of filters is set to 8.}
     \label{fig:cifar10-Fourier-filters}
\end{figure}

\begin{figure}
  \centering
    \includegraphics[width=0.5\textwidth]{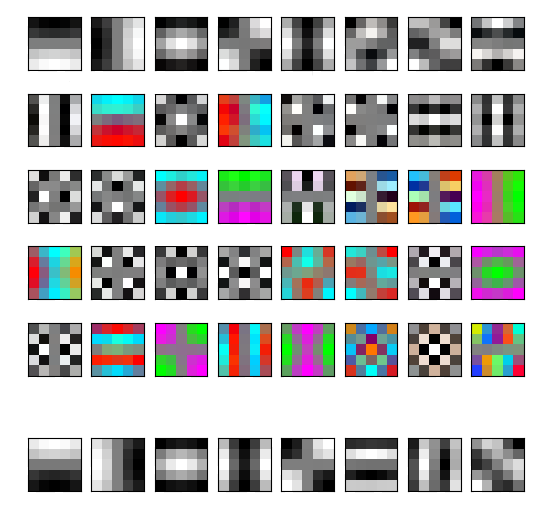}
    \caption{The PCA filters learned from \emph{CIFAR10} dataset. Top: the first stage filters, the number of filters for each channel is set to 40. Bottom: the second stage filters, the number of filters is set to 8.}
    \label{fig:cifar10-PCA-filters}
\end{figure}

We now look at the learned filters of FourierNet-2 and PCANet-2 from \emph{CIFAR10} dataset, which are shown in Figure \ref{fig:cifar10-Fourier-filters} and Figure \ref{fig:cifar10-PCA-filters} respectively. One could easily tell that the first stage filters of FourierNet-2 includes both low frequency and high frequency components, and the second stage filters consist of mainly low frequency filters, while one can hardly get intuitive sense from the visualization of PCANet-2 filters learned from \emph{CIFAR10} dataset.

\begin{table}[t]
\centering
\caption{Testing accuracy(\%) of two-stage models on CIFAR10}
\label{tab:accuracy-cifar10}
\begin{tabular}{ll}
\toprule
Model         & Accuracy \\ 
\midrule
FourierNet-2  & 67.70 \\
PCANet-2      & 70.95 \\
Fourier-PCA     & 68.30     \\
PCA-Fourier     & 69.75   \\ 
\bottomrule
\end{tabular}
\end{table}


\section{Conclusion}
In this paper, we propose adopting orthogonal basis in frequent domain like from Discrete Fourier Transformation or wavelets analysis as candidates for filter vectors in CNN. Different from \cite{chan2015pcanet}, our filter vectors are data independent, with no requirement for solving any optimization problem in the selection procedure, thereby rendering the whole process more transparent and understandable. Through extensive experiments, it is demonstrated that our method has witnessed comparable results in several benchmark data sets. Furthermore, analysis in frequency domain for computer vision can provide us with insights from different perspectives that cannot be achieved by principal component analysis. In the future, we plan to explore more datasets so as to figure out scenarios where these methods fit best and give full play to their effectiveness.

\clearpage 
\bibliographystyle{plainnat}   
{
\small
\bibliography{ijcai20}
}

\end{document}